\documentclass[10pt,twocolumn,letterpaper]{article}

\usepackage[pagenumbers]{cvpr} 
\usepackage{graphicx}
\usepackage{amsmath}
\usepackage{amssymb}
\usepackage{booktabs}
\usepackage{xcolor}         
\usepackage{graphicx}
\usepackage{siunitx}
\usepackage{caption}
\usepackage{subcaption}
\graphicspath{ {./imgs/} }
\newsavebox\CBox
\def\textBF#1{\sbox\CBox{#1}\resizebox{\wd\CBox}{\ht\CBox}{\textbf{#1}}}
\newcommand{\bestscore}[1]{\textcolor{red}{\textBF{#1}}}
\newcommand{\secondscore}[1]{\textcolor{blue}{\underline{#1}}}

\usepackage[pagebackref,breaklinks,colorlinks]{hyperref}
\usepackage[capitalize]{cleveref}
\crefname{section}{Sec.}{Secs.}
\Crefname{section}{Section}{Sections}
\Crefname{table}{Table}{Tables}
\crefname{table}{Tab.}{Tabs.}

\begin{document}

\title{Blur Interpolation Transformer for Real-World Motion from Blur}

\author{Zhihang Zhong$^{1,2}$\qquad
Mingdeng Cao$^{1}$\qquad
Xiang Ji$^{1,}$\qquad
Yinqiang Zheng$^{1}$\qquad
Imari Sato$^{1,2}$
\\
$^1$The University of Tokyo, Japan\qquad
$^2$National Institute of Informatics, Japan
\\
{\tt\small zhong@is.s.u-tokyo.ac.jp\qquad \{cmd,jixiang\}@g.ecc.u-tokyo.ac.jp}\\ {\tt\small yqzheng@ai.u-tokyo.ac.jp\qquad imarik@nii.ac.jp}
}

\maketitle

\begin{abstract}
   This paper studies the challenging problem of recovering motion from blur, also known as joint deblurring and interpolation or blur temporal super-resolution. The challenges are twofold: 1) the current methods still leave considerable room for improvement in terms of visual quality even on the synthetic dataset, and 2) poor generalization to real-world data. To this end, we propose a blur interpolation transformer (BiT) to effectively unravel the underlying temporal correlation encoded in blur. Based on multi-scale residual Swin transformer blocks, we introduce dual-end temporal supervision and temporally symmetric ensembling strategies to generate effective features for time-varying motion rendering. In addition, we design a hybrid camera system to collect the first real-world dataset of one-to-many blur-sharp video pairs. Experimental results show that BiT has a significant gain over the state-of-the-art methods on the public dataset Adobe240. Besides, the proposed real-world dataset effectively helps the model generalize well to real blurry scenarios. Code and data are available at \href{https://github.com/zzh-tech/BiT}{https://github.com/zzh-tech/BiT}.
\end{abstract}

\section{Introduction}
\label{sec:intro}
Aside from time-lapse photography, motion blur is usually one of the most undesirable artifacts during photo shooting. Many works have been devoted to studying how to recover sharp details from the blur, and great progress has been made. Recently, starting from Jin~\etal~\cite{jin2018learning}, the community has focused on the more challenging task of recovering high-frame-rate sharp videos from blurred images, which can be collectively termed joint deblurring and interpolation~\cite{shen2020blurry,shen2020video} or blur temporal super-resolution~\cite{rozumnyi2021defmo,rozumnyi2021shape,rozumnyi2022motion,oh2021demfi}. This joint task can serve various applications, such as video visual perception enhancement, slow motion generation~\cite{oh2021demfi}, and fast moving object analysis~\cite{rozumnyi2021defmo,rozumnyi2021shape,rozumnyi2022motion}. For brevity, we will refer to this task as blur interpolation.

Recent works~\cite{jin2019learning,gupta2020alanet,shen2020blurry} demonstrate that the joint approach outperforms schemes that cascade separate deblurring and video frame interpolation methods. Most joint approaches follow the center-frame interpolation pipeline, which means that they can only generate latent frames for middle moments in a recursive manner. DeMFI~\cite{oh2021demfi} breaks this constraint by combining self-induced feature-flow-based warping and pixel-flow-based warping to synthesize latent sharp frame at arbitrary time $t$. However, even on synthetic data, the performance of current methods is still far from satisfactory for human perception. We find that the potential temporal correlation in blur has been underutilized, which allows huge space for performance improvement of the blur interpolation algorithm. In addition, blur interpolation suffers from the generalization issue because there is no real-world dataset to support model training.

\begin{figure*}[!t]
	\centering
	\includegraphics[width=\linewidth]{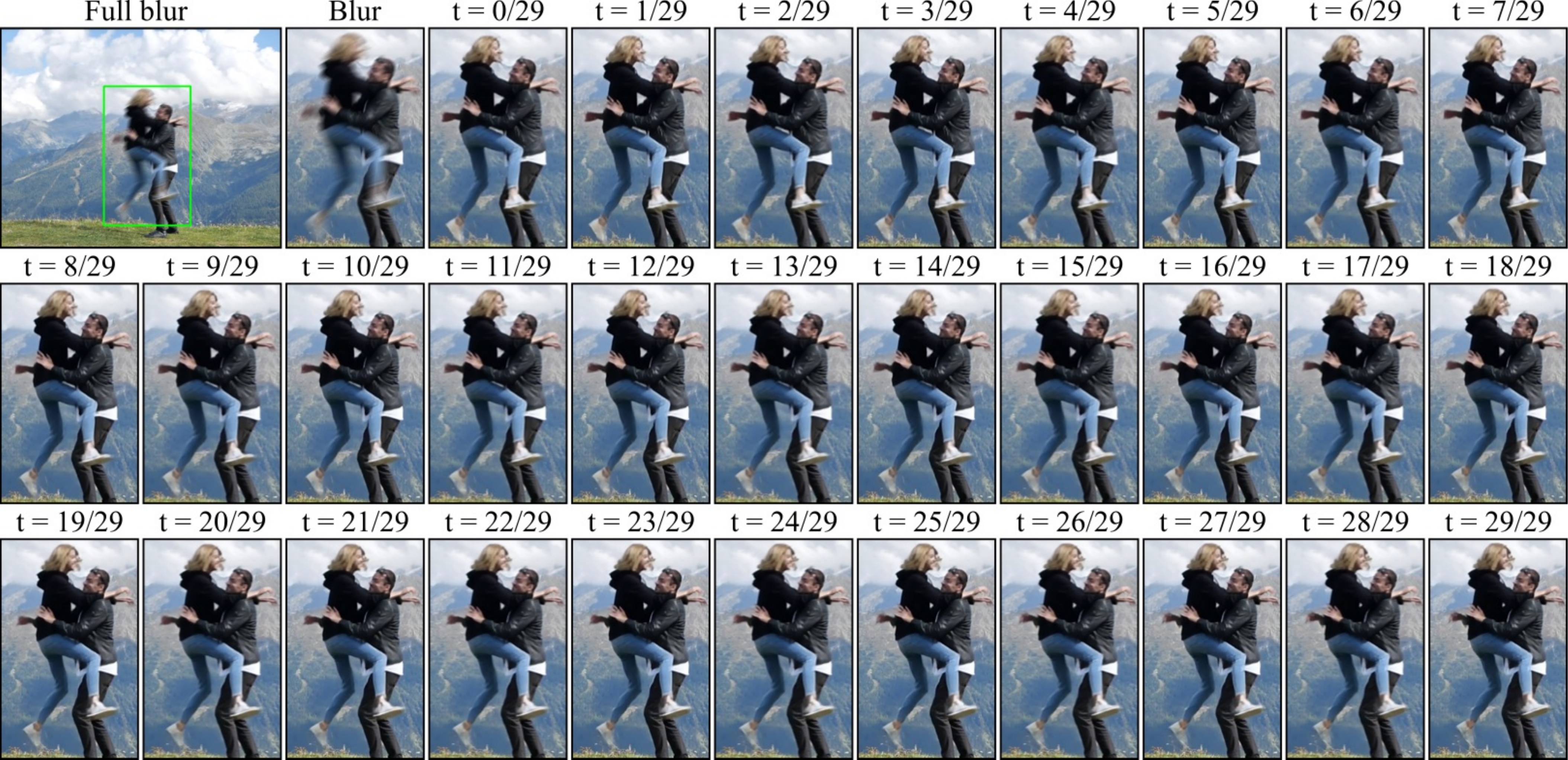}
	\caption{\textbf{Arbitrary blur interpolation by BiT.} This is an example of generating 30 sharp frames from blurred image using BiT.}
	\label{fig:arbitrary}
\end{figure*}

The goal of this work is to resolve the above two issues. In light of the complex distribution of time-dependent reconstruction and temporal symmetry property, we propose dual-end temporal supervision (DTS) and temporally symmetric ensembling (TSE) strategies to enhance the shared temporal features of blur interpolation transformer (BiT) for time-varying motion reconstruction. In addition, a multi-scale residual Swin transformer block (MS-RSTB) is introduced to empower the model with the ability to effectively handle the blur in different scales and to fuse information from adjacent frames. Due to our design, BiT achieves state-of-the-art on the public benchmark performance even without optical flow-based warping operations. Meanwhile, to provide a real-world benchmark to the community, we further design an accurate hybrid camera system following~\cite{rim2020real,zhong2020efficient} to capture a dataset (RBI) containing time-aligned low-frame-rate blurred and high-frame-rate sharp video pairs. Thanks to RBI, the real data generalization problem of blur interpolation can be greatly alleviated, and a more reasonable evaluation platform becomes available. With these improvements, our model presents impressive arbitrary blur interpolation performance, and we show an example of extracting 30 frames of sharp motion from the blurred image in Fig.~\ref{fig:arbitrary} for reference.

Our contributions can be summarized as follows:
\begin{itemize}
	\item We propose a novel transformer-based model, BiT, for arbitrary time motion from blur reconstruction. BiT outperforms prior art quantitatively and qualitatively with faster speed.
	\item We present and verify two successful strategies including dual-end temporal supervision and temporally symmetric ensembling to enhance the shared temporal features for arbitrary time motion reconstruction. 
	\item To the best of our knowledge, we provide the first real-world dataset for general blur interpolation tasks. We verify the validity of this real dataset and its meaningfulness to the community by extensive experiments.
\end{itemize}


\section{Related works}
\label{sec:related_works}
\subsection{General deblurring} 
\label{sec:deblurring}
The technological paradigm of general deblurring has experienced a shift from blur kernel estimation by traditional methods such as~\cite{levin2007blind,wulff2014modeling,ren2017video,hyun2015generalized} to direct sharp image regression by deep learning methods such as~\cite{nah2017deep,su2017deep,tao2018scale,wang2019edvr}. Various general network architectures, including CNNs, RNNs, and GANs have been explored in-depth for deblurring. Nah~\etal~\cite{nah2017deep} and Tao~\etal~\cite{tao2018scale} verify the effectiveness of the multi-scale (coarse-to-fine) CNNs structure for deblurring, while Zamir~\etal~\cite{zamir2021multi} prove the efficacy of multi-stage progressive strategy for deblurring.~\cite{hyun2017online,nah2019recurrent,zhou2019spatio,zhong2020efficient,wang2022efficient} customize their RNN structures to better exploit the long-term temporal correlation of blurry video. Wang~\etal~\cite{wang2019edvr} adopt deformable convolution~\cite{zhu2019deformable} to align the neighboring blurry frames to boost deblurring performance. Pan~\etal~\cite{pan2020cascaded} and Son~\etal~\cite{son2021recurrent} explicitly utilize optical flow for more accurate motion compensation. Moreover, GANs are explored by~\cite{kupyn2018deblurgan, kupyn2019deblurgan} to deblur images with the goal of better human perception. Recently, transformer~\cite{liang2021swinir} has made a splash in the low-level vision tasks. Restormer~\cite{zamir2022restormer}, RVRT~\cite{liang2022recurrent}, and VDTR~\cite{cao2022vdtr} are proposed to demonstrate the great performance of transformer structures in the general deblurring tasks.

\subsection{Blur interpolation}
\label{sec:blur_interpolation}
The aim of blur interpolation goes beyond traditional deblurring, which focuses on a one-to-one mapping between blurred and sharp images. Instead, it involves utilizing the temporal information present in motion blur to reconstruct a motion sequence. There are similar tasks that utilize the partial temporal information in rolling shutter distortion to extract video clips, such as~\cite{zhong2022bringing} and~\cite{fan2021inverting}. Jin~\etal~\cite{jin2018learning} are the first to exploit blur interpolation, extracting a sharp video clip from only one single blurry image. However, blur interpolation from singe image faces the fundamental problem of directional ambiguity. Considering the freedom of each individual and uniform blurred region, the solution space of blur interpolation will be exponential. Therefore, Jin~\etal propose a pairwise order-invariant loss to alleviate the fundamental directional ambiguity and help the model converge to a single solution. Then, Purohit~\etal~\cite{purohit2019bringing} utilize a motion representation, which is learned from videos by a self-supervised strategy, to further tackle the directional ambiguity. After that, Argaw~\etal~\cite{argaw2021restoration} leverage a spatial transformer network with multiple independent branches and a transformation consistency loss to simultaneously estimate the motion of middle time and other times within the exposure time. Zhong~\etal~\cite{zhong2022animation} are the first to explicitly account for such directional ambiguity by introducing a motion guidance representation. The motion guidelines enable their approach to produce multiple plausible solutions from the same blurred image, rather than just one as was the case before.

Taking blurry video as input~\cite{jin2019learning,shen2020blurry,shen2020video,argaw2021motion,oh2021demfi} for blur interpolation, directional ambiguity can be largely avoided based on the motion cues of adjacent frames. Specifically, Jin~\etal~\cite{jin2019learning} present a cascaded scheme of deblurring-first and interpolation-later for this setting. To mitigate the accumulated errors introduced in the cascaded scheme, Shen~\etal~\cite{shen2020blurry,shen2020video} propose a pyramid recurrent framework to estimate the latent sharp sequence without explicitly distinguishing the deblurring stage and interpolation stage. Argaw~\etal~\cite{argaw2021motion} implement blur interpolation by initially estimating the optical flow, and then predicting a motion sequence by warping the decoded features to the corresponding time points. Recently, Oh~\etal~\cite{oh2021demfi} propose DeMFI framework, which combine flow-guided attentive-correlation-based feature bolstering module and recursive boosting techniques to convert lower-frame-rate blurred videos to higher-frame-rate sharp videos with state-of-the-art performance. There are also some works specialized to implement blur interpolation for fast-moving objects such as~\cite{rozumnyi2021defmo,rozumnyi2021shape,rozumnyi2022motion}. Given a pre-estimated background and a blurred image with a fast-moving object, they project the object representation to a latent space, and can render the object to a specified time tick within the exposure time. While Pan~\etal~\cite{pan2019bringing} and Lin~\etal~\cite{lin2020learning} use an additional event camera as an aid to accomplish this task. Our approach further pushes the performance of this task on generic scenarios with dual-end temporal supervision and temporally symmetric ensembling strategies as well as a stronger backbone.

\subsection{Deblurring dataset}
\label{sec:deblur_dataset}
In the early stage, the research community applied various blur kernels to synthesize blurred images with uniform motion, such as~\cite{levin2009understanding,schmidt2013discriminative,lai2016comparative,shen2018deep}. A blurred image $I_b$ can be described as the convolution between a sharp image $I_s$ and a blurred kernel $K$ with optional Gaussian noise $N$:
\begin{equation}
	I_b = K * I_s + N.
\end{equation}
Then, to deal with more realistic situation with spatially varying blur, researchers adopt a scheme of averaging consecutive frames of a high-frame-rate sharp video to synthesize blurred and sharp image/video pairs. Based on this basic pipeline, Su~\etal~\cite{su2017deep} synthesize a dataset named DVD and additionally interpolates between sharp frames using optical flow to reduce ghosting artifacts in the synthesized blurry video. While Nah~\etal~\cite{nah2017deep} synthesize a dataset named GOPRO by applying an inverse gamma correction before averaging to reduce the effect of nonlinear transformations. Later, Nah~\etal~\cite{nah2019ntire} combine the strengths of DVD and GOPRO to create a larger and more diverse synthetic deblurring dataset dubbed REDS. Regarding the more challenging blur interpolation task, previous works like~\cite{jin2019learning,shen2020blurry,shen2020video,argaw2021motion,oh2021demfi} also use discrete frames to create datasets of one-to-many blur-sharp pairs.

Models trained on synthetic data suffer from the persistent problem of being difficult to generalize to real-world data. Thus, many researchers have started to use hybrid camera systems to collect real-world datasets for low-level vision tasks~\cite{zhong2020efficient,zhong2021towards,cao2022learning,rim2020real}. Rim~\etal~\cite{rim2020real} and Zhong~\etal~\cite{zhong2020efficient,zhong2021towards} build hybrid camera systems based on beam-splitter to collect real image deblurring dataset (RealBlur) and real video deblurring datasets (BSD and BS-RSCD), respectively. Inspired by the success of real-world datasets, we customize a hybrid system to collect a real dataset (RBI) of time-aligned low-frame-rate and high-frame-rate blur-sharp video pairs. We believe RBI can benefit the community to better benchmark blur interpolation algorithms.

\section{Blur interpolation transformer}
\label{sec:bit}

\begin{figure*}[!t]
	\centering
	\includegraphics[width=.95\linewidth]{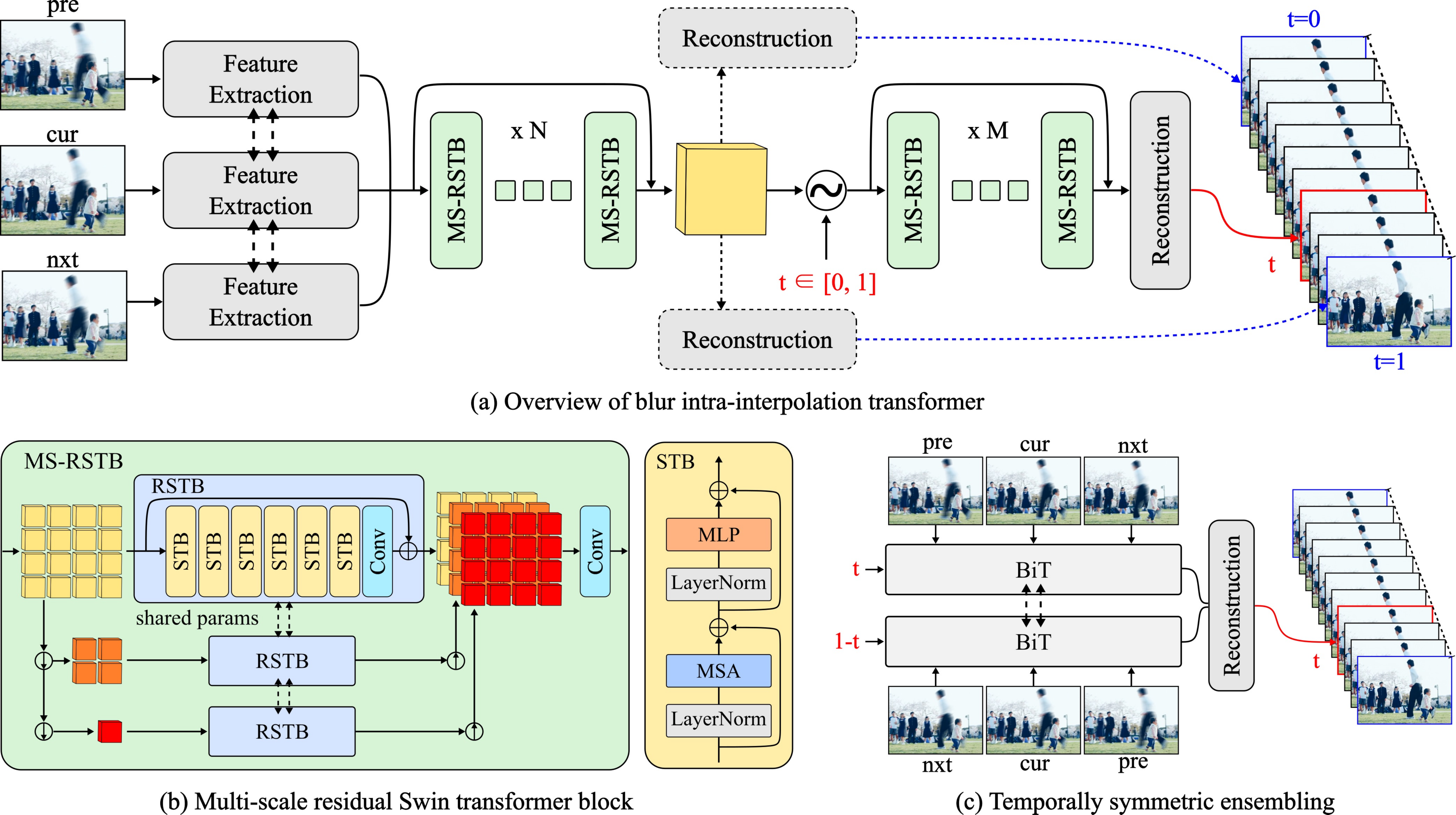}
	\caption{\textbf{Overview of BiT.} (a) is the proposed blur interpolation transformer that takes three consecutive blurry images as inputs to generate shared temporal features. Then, shared features and a normalized t are incorporated to render a sharp motion for a specified moment in the exposure time. Besides, there is an additional reconstruction layer with twice as many channels as the final reconstruction layer for reconstructing the results from the shared features for dual-end time points. (b) is the details of multi-scale residual Swin transformer block. A RSTB module~\cite{liang2021swinir} is shared to tackle blur and fuse neighboring features in a coarse-to-fine manner. (c) is temporally symmetric ensembling strategy. A pre-trained BiT with a new reconstruction layer is adopted to fuse the reconstruction features in forward (t) and reverse (1-t) orders.}
	\label{fig:model}
\end{figure*}

The overview architecture of our blur interpolating transformer (BiT) is shown in Fig.~\ref{fig:model} (a). BiT focuses on interpolating a sharp motion $\hat{I}_s^t$ for a blurred image given an arbitrary $t$ during the exposure time. Regarding the directional ambiguity in this problem, considering that modern cameras have short exposure times and that the relative speed between camera and scene is not very fast, we follow the implicit assumption of previous works~\cite{jin2019learning,shen2020video,oh2021demfi,zhong2022animation} that there is no ambiguity when the input is video. Thus, we also use neighboring frames as auxiliary inputs to get rid of this issue. The inference process of the target model $\mathcal{F}$ can be described as follows:
\begin{equation}
	\label{eq:inference}
	\hat{I}_s^t = \mathcal{F}\left(\mathbf{I_b}, t\right),
\end{equation}
where $\mathbf{I_b}=\left\{I_b^{pre}, I_b^{cur}, I_b^{nxt}\right\}$, denoting the input set of previous, current, and next blurred images. $t$ represents a specific time point during exposure of $I_b^{cur}$ with a normalized value range of $t\in[0, 1]$. Apart from the lightweight reconstruction layer $\mathcal{F}_R$, the model is divided into two stages, including a shared temporal feature extraction stage $\mathcal{F}_N$ and an arbitrary motion rendering stage $\mathcal{F}_M$. $\mathcal{F}_N$ consists of a shared down-sampling convolutional block for shallow feature extraction and followed by $N$ blocks of multi-scale residual Swin transformer blocks (MS-RSTB). The $\mathcal{F}_M$ consists of $t$ encoding module and $M$ MS-RSTBs. Then, the inference process Eq.~\ref{eq:inference} can be reformulated as:
\begin{equation}
	\hat{I}_s^t = \mathcal{F_R}\left(\mathcal{F_M}\left(\mathcal{F_N}\left(\mathbf{I_b}\right), t\right)\right).
\end{equation}
The fact that only the latter part $\mathcal{F}_M$ needs to be repeated when performing multiple time inferences can optimize the network multiplexing efficiency. Extracting well-formed shared temporal features is the key to achieving arbitrary motion rendering under discrete-time supervision. We then present the module and training strategies introduced for improving the performance of arbitrary motion rendering from blur, one by one.

\paragraph{Multi-scale residual Swin transformer block.} Inspired by the powerful modelling ability of residual Swin transformer block (RSTB)~\cite{liang2021swinir}~\footnote{Please also refer to Appendix for more details.} for image restoration task, a new and efficient backbone block is constructed by introducing the classic coarse-to-fine multi-scale structure. The proposed MS-RSTB reuses one RSTB to process interpolated features at different scales, and then a convolutional layer is used to fuse the features to the same shape as the input features, as illustrated in Fig.~\ref{fig:model} (b). The original input feature of shape $\mathbb{R}^{C\times H\times W}$ is interpolated to the shape $\mathbb{R}^{C\times H/r^{s-1}\times W/r^{s-1}}$ regarding to the scale level $s \in \{1, \cdots, S\}$ and the rescale ratio $r$. The computational complexity of MS-RSTB is increased as follows:
\begin{equation}
	\Omega\left(\text{MS-RSTB}\right) \approx \frac{1-\left(1/r\right)^{2S}}{1-\left(1/r\right)^{2}} \Omega\left(\text{RSTB}\right),
\end{equation}
where we set $S=3$ and $r=2$ so that there is an additional computational cost less than $1/3$. Since the window size of RSTB is fixed, the multi-scale features allow the self-attentive mechanism to be applied from global to local. This facilitates the handling of different scales of blur and the fusion of informative features from adjacent frames without prior knowledge of the range of motion.

\begin{figure*}[!t]
	\centering
	\includegraphics[width=\linewidth]{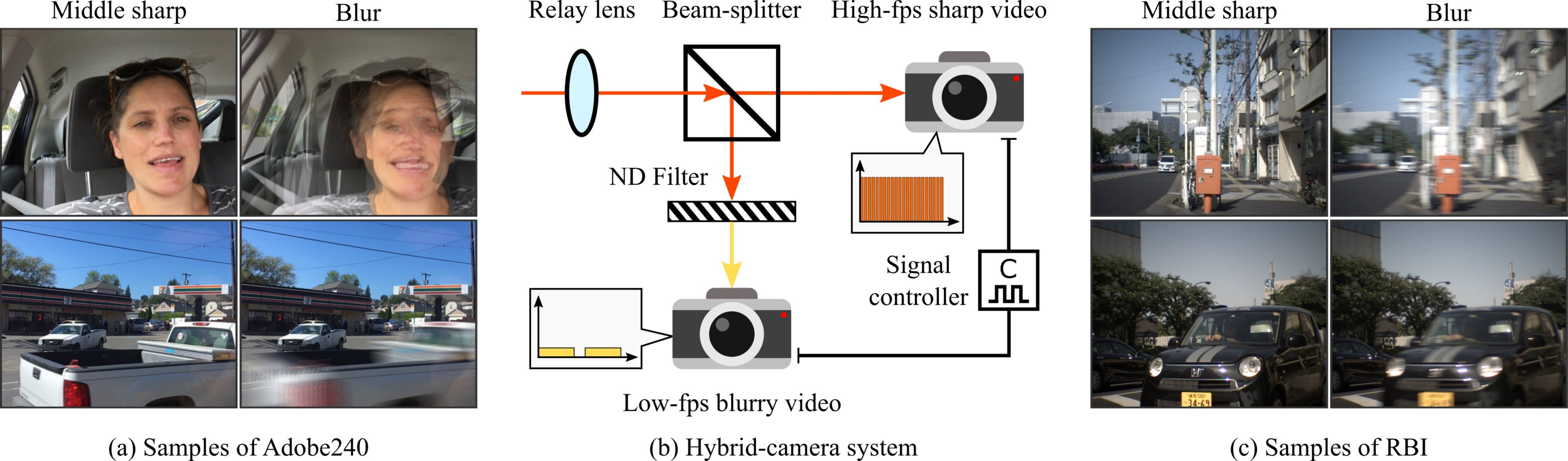}
	\caption{\textbf{Real blur interpolation dataset.} (a) are samples from synthetic dataset Adobe240 with unnatural spikes or steps in the blur trajectory. (b) is the schematic of our hybrid camera system. (c) are samples from our real-world dataset RBI with natural blur.}
	\label{fig:dataset}
\end{figure*}

\paragraph{Dual-end temporal supervision.} A key observation for time-varying motion rendering from blur is that the difficulty increases from the middle moment to both sides with greater temporal difference. One insight is that if, without any temporal cues, the model is able to extract qualified features to render the most extreme moments of motion, such learned features are well-formed in respect to the varying $t$ to better render motions at other moments. Thus, we propose a simple yet effective learning strategy, called dual-end temporal supervision (DTS), to underpin and spread the shared temporal features. Specifically, the shared temporal features,~\ie, the yellow cube in Fig.~\ref{fig:model} (a), are forced to restore the motions of the two end time points using an additional lightweight reconstruction layer $\mathcal{F^{D}_R}$ without any $t$ encoding:
\begin{equation}
	\left\{\hat{I}_s^t\ |\ t=0,1\right\} = \mathcal{F^{D}_R}\left(\mathcal{F_N}\left(\mathbf{I_b}\right)\right).
\end{equation}
Note that this extra reconstruction layer will be discarded in the test mode. DTS acts as anchors for the boundaries, making the shared temporal features more conducive to motion rendering in a continuous-time manner.

\paragraph{Temporally symmetric ensembling.} Another insight into the arbitrary time motion rendering from blur arises from the consistency of the results from temporally forward and reverse inputs. Intuitively, the rendered motion at $t$ of $\mathbf{I_b}$ can also be represented as the rendered motion at $1-t$ of the temporally inverse blurred inputs $\mathbf{I_b^{inv}}=\left\{I_b^{nxt}, I_b^{cur}, I_b^{pre}\right\}$. Thus, given a pre-trained BiT, we can further enhance it by fusing forward and inverse complementary features, named temporally symmetric ensembling (TSE). As shown in Fig.~\ref{fig:model} (c), during the fine-tuning process, the last reconstruction layer is replaced by a new layer $\mathcal{F^{T}_{R}}$ that can accept twice the number of input channels. The inference process with TSE strategy is as follows:
\begin{equation}
	\hat{I}_s^t = \mathcal{F^{T}_R}\left(\mathcal{F_M}\left(\mathcal{F_N}\left(\mathbf{I_b}\right), t\right),\  \mathcal{F_M}\left(\mathcal{F_N}\left(\mathbf{I_b^{inv}}\right), 1-t\right)\right).
\end{equation}

Thanks to the proposed MS-RSTB and the designed temporal optimization strategies, taking the L1 loss to supervise reconstruction results is sufficient to ensure an excellent performance of the model. The total loss terms are as follows:
\begin{equation}
	\mathcal{L} = \mathcal{L}_1\left(\hat{I}_s^t, I_s^t\right) + \lambda\left(\mathcal{L}_1\left(\hat{I}_s^0, I_s^0\right) + \mathcal{L}_1\left(\hat{I}_s^1, I_s^1\right)\right).
\end{equation}
In the training phase, $t$ is only randomly sampled from the available discrete time points of the corresponding dataset. However, in the test phase, $t$ can take any continuous value between 0 and 1. As for the $t$ encoding, we expand the spatial size of $t$ to the same size as the shared features. Then, we merge it into the channel dimension of the shared features, followed by a linear layer for feature fusion. Empirically, we find that such simple encoding can provide good performance, slightly better than the widely used frequency encoding such as~\cite{mildenhall2020nerf}.


\section{Real-world blur interpolation dataset}
\label{sec:dataset}

\begin{figure*}[!t]
	\centering
	\includegraphics[width=\linewidth]{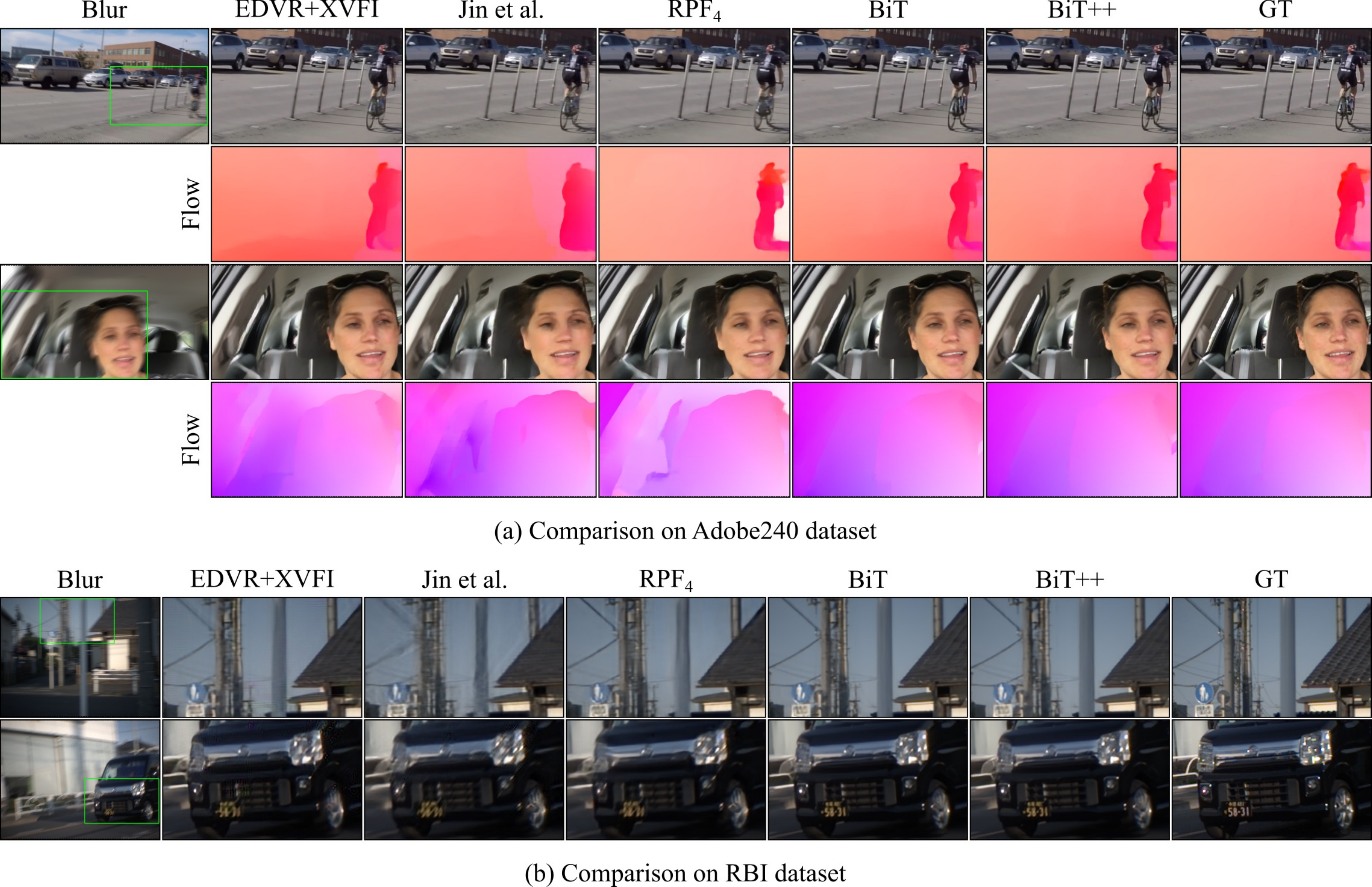}
	\caption{\textbf{Qualitative comparisons on Adobe240 dataset and real-world dataset RBI.}}
	\label{fig:quality}
\end{figure*}

\begin{figure*}[!t]
	\centering
	\includegraphics[width=\linewidth]{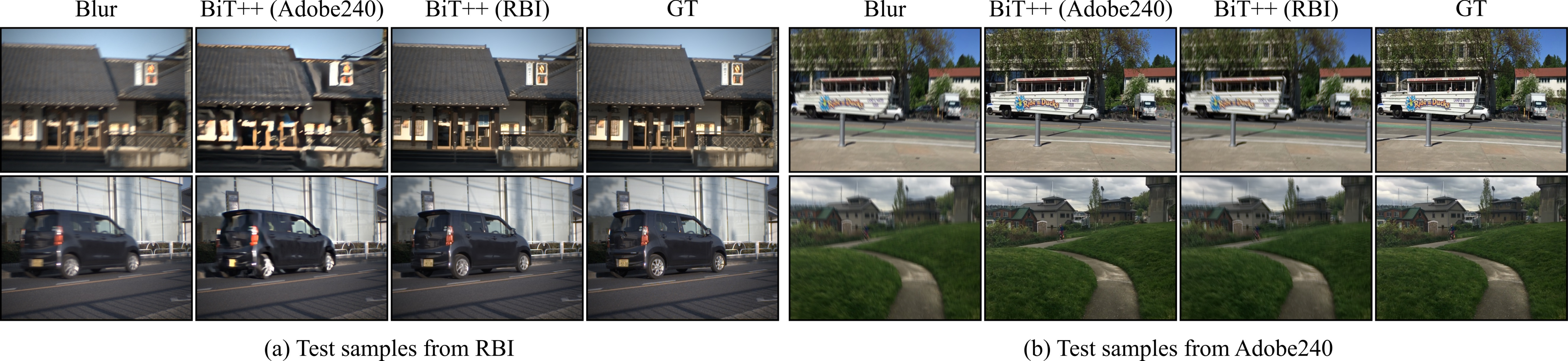}
	\caption{\textbf{Cross-validation between synthetic dataset Adobe240~\cite{shen2020video} and real-world dataset RBI.} BiT++(Adobe240) and BiT++(RBI) represent the model trained on Adobe240 and RBI. The model trained on synthetic data will cause artifacts when tested on real data.}
	\label{fig:cross}
\end{figure*}

\begin{table*}[!t]
	\setlength{\tabcolsep}{5pt}
	\caption{\textbf{Comparison with the state-of-the-arts on synthetic dataset Adobe240 and our real-world dataset RBI.} \bestscore{Red} denotes the best performance, and \secondscore{blue} denotes the second best performance. Runtime is calculated uniformly using images from the Adobe240 dataset with size of $640\times 352$ on a single RTX2080 Ti GPU.}
	\label{table:comparison}
	\centering
	\setlength{\tabcolsep}{9pt}
	\begin{tabular}{lccccccc}
		\toprule
		& \multicolumn{2}{c}{Adobe240} & \multicolumn{2}{c}{RBI} & \multicolumn{2}{c}{Runtime}  & \\
		\cmidrule(r){2-3}
		\cmidrule(r){4-5}
		\cmidrule(r){6-7}
		& PSNR $\uparrow$ & SSIM $\uparrow$ & PSNR $\uparrow$ & SSIM $\uparrow$ & 1x [s] $\downarrow$ & 60x [s] $\downarrow$ & Params [M] $\downarrow$\\
		\midrule
		EDVR~\cite{wang2019edvr}+XVFI~\cite{sim2021xvfi} & 33.19 & 0.934 & 28.17 & 0.847 & 0.294 & 17.64 & 29.2\\
		Jin~\etal~\cite{jin2019learning} & 32.47 & 0.924 & 27.73 & 0.853 & \secondscore{0.250} & 15.00 & \secondscore{10.8}\\
		RPF$_4$~\cite{shen2020video}& 33.32 & 0.935 & 28.55 & 0.872 & 0.746 & 44.76 & 11.4\\
		DeMFI~\cite{oh2021demfi} & \secondscore{34.34} & 0.945 & 29.03 & 0.895  & 0.513 & 30.78 & \bestscore{7.41} \\
		\midrule
		BiT & \secondscore{34.34} & \secondscore{0.948} & \secondscore{29.90} & \secondscore{0.900} & \bestscore{0.203} & \bestscore{5.76} & 11.3\\
		BiT++ & \bestscore{34.97} & \bestscore{0.954} & \bestscore{30.45} & \bestscore{0.908} & 0.395 & \secondscore{11.64} & 11.3\\
		\bottomrule
	\end{tabular}
\end{table*}

\paragraph{Limitation of synthetic dataset.} First, let us briefly review the synthetic pipeline of previous works. Taking Adobe240 from~\cite{shen2020blurry,shen2020video} as an example, a sliding window with size of $M$ frames is used to average the sharp images. The blurred image can be generated as follows:
\begin{equation}
	I_{b} = \frac{1}{M} \sum_{m=1}^M\left(I_s^m\right),
\end{equation}
where $M=11$. The samples of Adobe240 are illustrated in Fig.~\ref{fig:dataset} (a). We can observe the unnatural spikes or steps in the blur trajectory due to the discrete averaging process. Therefore, synthetic datasets like Adobe240 may not reflect the actual difficulty of the blur interpolation task. In addition, the gap between synthetic blur and real blur may lead to generalization problems for models trained on synthetic datasets.

\paragraph{Hybrid camera system.} Building a real-world dataset for the blur interpolation task becomes an urgent need. Blur should occur naturally in the form of signal accumulation, as follows:
\begin{equation}
	I_{b} = \int_{0}^{\tau}S(t)dt,
\end{equation}
where $\tau$ denotes the exposure time, and $S(t)$ denotes the signal captured by the camera sensor at time $t$. To this end, we design a hybrid camera system as illustrated in Fig.~\ref{fig:dataset} (b). Specifically, two BITRAN CS-700C cameras are physically aligned to the beam-splitter by laser calibration. During shooting, the light is split in half and goes into the camera with high and low frame rate modes. The low-frame-rate camera adopts long exposure scheme to capture the blurred video. Besides, a ND filter with about 10\% transmittance is installed before the low-frame-rate camera to ensure photometric balance between the blurred frames and the corresponding sharp frames from the high-frame-rate camera.

\paragraph{RBI dataset.} We use this customized hybrid camera system to collect 55 video pairs as real-world blur interpolation (RBI) dataset. The frame-rate of blurred video and the corresponding sharp video are \SI{25}{fps} and \SI{500}{fps}, respectively. The exposure time of blurred image is \SI{18}{ms}, while the exposure time of sharp image is nearly \SI{2}{ms}. This means that there are 9 sharp frames corresponding to one blurred frame, and 11 sharp frames exist in the readout deadtime between adjacent blurred frames. The image size is $640\times480$. We shoot videos of normal urban scenes with various motion modes, including ego-motion, object motion, and both.  In addition to blur interpolation, RBI can serve as a dataset to measure the performance of blur synthesis algorithms, such as~\cite{brooks2019learning}.

\section{Experiments}
\label{sec:experiments}

\begin{figure*}[!t]
	\centering
	\includegraphics[width=\linewidth]{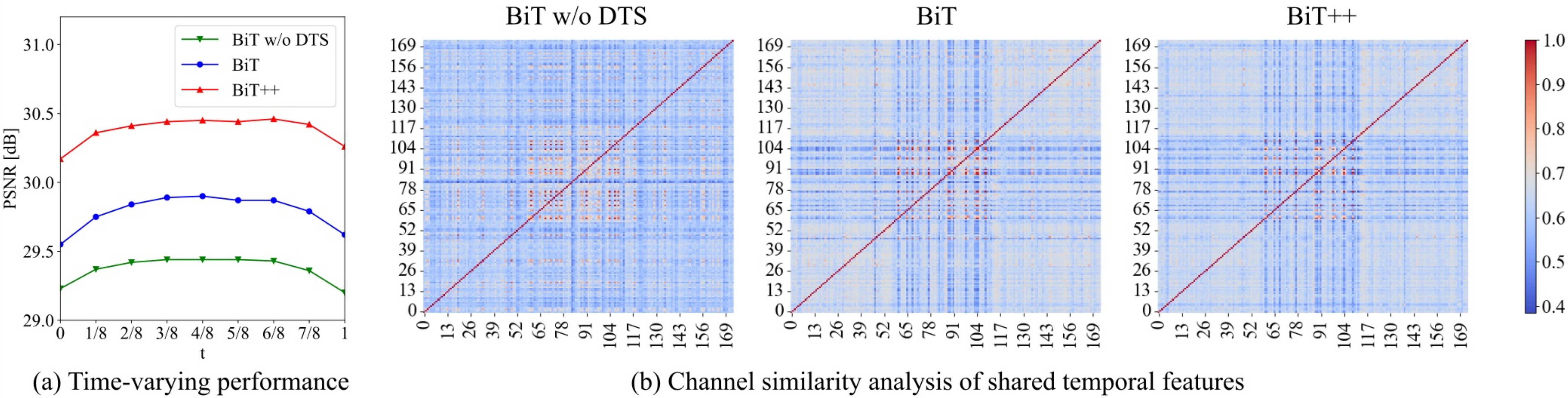}
	\caption{\textbf{Ablation studies for temporal strategies.} (a) shows time-varying performance. (b) is CKA analysis of shared temporal features. This visualization is created based on the test data of RBI.}
	\label{fig:ablation}
\end{figure*}

\begin{table*}[!t]
	\caption{\textbf{Ablation studies.} BiT w/o MS denotes BiT using single-scale RSTB module. BiT w/o DTS denotes BiT without dual-end temporal supervision. BiT+ denotes BiT that has the same training epochs as BiT++. 
	}
	\label{table:ablation}
	\centering
	\setlength{\tabcolsep}{5pt}
	\begin{tabular}{lcccccccccc}
		\toprule
		& \multicolumn{5}{c}{Adobe240} & \multicolumn{5}{c}{RBI} \\
		\cmidrule(r){2-6}
		\cmidrule(r){7-11}
		& BiT w/o MS & BiT w/o DTS & BiT & BiT+ & BiT++ & BiT w/o MS & BiT w/o DTS & BiT & BiT+ & BiT++\\
		\midrule
		PSNR $\uparrow$ & 33.96 & 34.10 & 34.34 & \secondscore{34.52} & \bestscore{34.97} & 29.40 & 29.44 & 29.90 & \secondscore{29.99} & \bestscore{30.45} \\
		SSIM $\uparrow$ & 0.944 & 0.946 & \secondscore{0.948} & 0.946 & \bestscore{0.954} & 0.893 & 0.894 & 0.900 & \secondscore{0.901} & \bestscore{0.908} \\
		\bottomrule
	\end{tabular}
\end{table*}

We optimize the loss using AdamW~\cite{loshchilov2017decoupled} in the PyTorch framework~\cite{paszke2019pytorch}. $\lambda$ is empirically set as 0.5. In both the initial training or fine-tuning phase, the learning rate is scheduled by the cosine scheduler from $1\times10^{-4}$ to $1\times10^{-6}$. Common data augmentation operations, including flipping, rotation, and cropping of size $256\times 256$, are used. We set $M=3$ and $N=3$ as the default BiT settings for $\mathcal{F_N}$ and $\mathcal{F_M}$. The number of heads and channel size of self-attention is set to 6 and 174. Regarding the partitioning of the dataset, Adobe240 is the same as previous work~\cite{shen2020blurry,shen2020video}; while 50 videos of RBI are used for training and the remaining 5 videos are used for testing. We train BiT on Adobe240 with a batch size of 32 for 800 epochs, and finetune it with TSE strategy for another 400 epochs, on 8 NVIDIA Tesla V100 GPUs. Since the size of RBI is smaller than Adobe240, we double the number of epochs and reduce the batch to 8 to get more iterations for training. Regarding the other models, we retrain them on each dataset for a fair comparison. In addition to this section, we encourage readers to refer to the appendices for more details regarding the proposed RBI dataset, as well as supplementary ablation studies and experiments.

\paragraph{Quantitative and qualitative results.} We compare our method with previous state-of-the-arts including Jin~\etal~\cite{jin2019learning}, RPF$_4$~\cite{shen2020video}, DeMFI~\cite{oh2021demfi}, and a cascaded method consists of deblurring model EDVR~\cite{wang2019edvr} and interpolation model XVFI~\cite{sim2021xvfi}. Since the Adobe240 dataset has no readout deadtime, we follow the 2x temporal super-resolution setting of this dataset to compare with other methods. While on the RBI dataset, we compare the middle deblurred images, because in the real case, there is a readout deadtime.

Quantitative results are shown in Table~\ref{table:comparison}. We name our model with the TSE strategy as BiT++ and the one without is BiT.
BiT++ can outperform the prior art on Adobe240 and on RBI by a large margin. 
Besides, the more time points are derived from the same input, the faster our model becomes. BiT achieves 60 inferences in 5.76 seconds, while maintaining favorable performance. Qualitative results are shown in Fig.~\ref{fig:quality}. We can see that the predictions of BiT and BiT++ are closer to the groundtruth with clearer details on both Adobe240 and RBI. We further utilize RAFT~\cite{teed2020raft} to estimate the optical flow between two adjacent predicted frames on Adobe240, as illustrated in Fig.~\ref{fig:quality} (a). The optical flow of our results is also closer to the groundtruth, which indicates better motion consistency.

\begin{table*}[!t]
	\setlength{\tabcolsep}{13pt}
	\caption{\textbf{Effect of \# of MS-RSTB.} The performance is evaluated on Adobe240 using BiT.}
	\label{table:num_msrstb}
	\centering
	\begin{tabular}{lccccccc}
		\toprule
		& $N=0$ & $N=1$ & $N=2$ & $N=3$ & $N=4$ & $N=5$ & $N=6$\\
		& $M=6$ & $M=5$ & $M=4$ & $M=3$ & $M=2$ & $M=1$ & $M=0$\\
		\midrule
		PSNR $\uparrow$ & 34.08 & 34.09 & 34.18 & \bestscore{34.34} & \secondscore{34.30} & 34.05 & 27.13\\ 
		SSIM $\uparrow$ & \secondscore{0.947} & 0.942 & 0.943 & \bestscore{0.948} & \bestscore{0.948} & 0.944 & 0.832\\
		60x Runtime [s] $\downarrow$ & 11.34 & 9.36 & 7.98 & 5.76 & 4.02 & \secondscore{2.16} & \bestscore{0.36} \\
		\bottomrule
	\end{tabular}
\end{table*}

\paragraph{Dataset cross-evaluation.} To demonstrate the need for a real dataset, we conduct experiments on cross-evaluation between the synthetic dataset Adobe240 and the real-world dataset RBI. First, we show the results of RBI samples predicted by independent BiT++ models trained on Adobe240 and RBI in Fig.~\ref{fig:cross} (a). We can observe severe artifacts in the results of the model trained on Adobe240. Conversely, testing on synthetic data shown in Fig.~\ref{fig:cross} (b), we find that the model trained on RBI does not introduce artifacts, even if it could not remove the synthetic blur. This experiment demonstrates the risks of training a model on a synthetic dataset for the blur interpolation task,  which is consistent with the findings of previous work~\cite{zhong2022real}.

\paragraph{Ablation studies.}
In order to verify the validity of the proposed new module and strategies, we perform relevant ablation experiments. We show the results of BiT with only single-scale RSTB (denoted as BiT w/o MS), BiT without DTS (denoted as BiT w/o DTS), and BiT with the same total training epochs as BiT++ (denoted as BiT+) in Table.~\ref{table:ablation}. The MS-RSTB, DTS, and TSE can bring 0.38dB, 0.24dB, and 0.45dB gain on Adobe240, as well as 0.50dB, 0.46dB, and 0.46dB gain on RBI, respectively. We also present the curves of time-varying performance of ablated models on RBI, as illustrated in~Fig.\ref{fig:ablation} (a). The full model BiT++ improves the performance of all time points within the exposure time. 

To further explain the benefits from our temporal feature enhancement strategies, we use the central kernel alignment (CKA)~\cite{kornblith2019similarity} to measure the channel-wise similarity of the extracted shared temporal features, as shown in Fig.~\ref{fig:ablation} (b). The modified CKA calculation process is as follows:
\begin{equation}
	\label{eq:cka_map}
	\text{CKA}(i, j) = \frac{\text{HSIC}(G(F_i), G(F_j))}{\sqrt{\text{HSIC}(G(F_i), G(F_i))\text{HSIC}(G(F_j), G(F_j))}},
\end{equation}
where $i$ and $j$ are the channel indices of extracted shared temporal feature $F\in \mathbb{R}^{C\times H \times D}$. $\text{HSIC}$ and $G$ are the functions to calculate the Hilbert-Schmidt independence criterion and Gram matrices, respectively. We find that after applying the temporal feature enhancement strategy including DTS and TSE, the shared features show a significant functional stratification in the channel dimension. In particular, the channel features of BiT++ are clustered into three rectangular blocks. We speculate that the first and third feature blocks aggregate common features shared by different time inferences, while the middle feature block aggregates more differentiated features ready to be retrieved accordingly based on the given time query.

In addition, we analyze how to distribute the number of MS-RSTB before and after the shared temporal features in Table~\ref{table:num_msrstb}. To keep the total number of parameters constant, the total number of MS-RSTB is set to 6. The MS-RSTBs before the shared feature are shared at different moments of inference. When inferring the results for multiple time points, the temporal features obtained from the N MS-RSTB blocks need to be computed only once. Thus, larger $N$ makes multiple inferences faster. We find that expanding $N$ to 4 or 5 results in a significant speedup, but only a slight performance loss compared to the default setting $N=3$.


\section{Conclusion, limitation, and future work}
\label{sec:conclusion}
We propose a novel and efficient model, BiT, to realize arbitrary time blur interpolation with state-of-the-art performance. In addition, we present a real-world dataset RBI that enables the first real-world benchmark for blur interpolation task. However, current limited discrete supervision may not be sufficient to cope with very fast motions. Besides, to better cope with different real-world situations, our dataset needs to be expanded to include video pairs with different devices and different exposure parameters. We believe that the reversed process, \ie, learning to synthesize real blur using successive sharp frames from RBI, is also an interesting and valuable direction for the future.

\section*{Acknowledgement}
\label{sec:acknowledgement}
This work was supported by JST, the establishment of university fellowships towards the creation of science technology innovation, Grant Number JPMJFS2108.

{\small
\bibliographystyle{ieee_fullname}
\bibliography{egbib}
}

\clearpage
\appendix
\section{Supplementary details}
\begin{figure}[!t]
\centering
    \includegraphics[width=\linewidth]{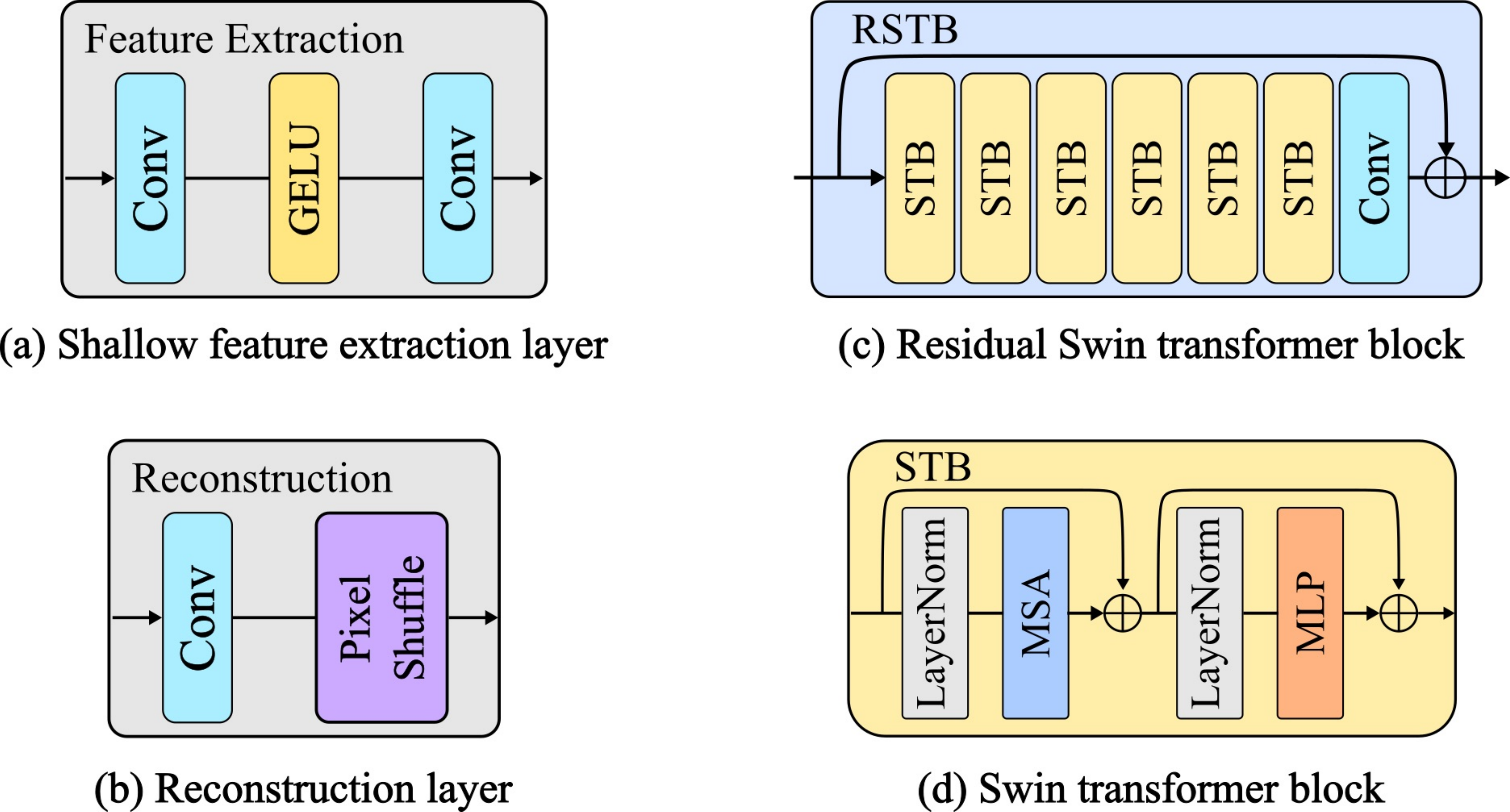}
    \caption{\textbf{Structure of sub-modules.} (a) is the structure of shared shallow feature extraction layer. (b) is the structure of reconstruction layer. (c) is the structure of residual Swin transformer block. (d) is the structure of Swin transformer block.}
    \label{fig:modules}
\end{figure}

\paragraph{Blur interpolation transformer (BiT).} First, we add some details of the BiT network structure in this part. The shared shallow feature extraction layer is illustrated in Fig.~\ref{fig:modules} (a). It consists of two $3\times3$ 2d convolution layers with stride equal to 2, and a GELU~\cite{hendrycks2016gaussian} activation layer between them. The lightweiht reconstruction layer is shown in Fig.~\ref{fig:modules} (b). It only consists of a $3\times3$ 2d convolution layer and a PixelShuffle~\cite{shi2016real} layer with upscale factor equal to 4. Regarding residual Swin transformer block (RSTB), we borrow the structure from SwinIR~\cite{liang2021swinir}, as illustrated in Fig.~\ref{fig:modules} (c). It consists of 6 stacked Swin transformer blocks (STB) and a $3\times3$ 2d convolution layer at the end for learning the features in a residual manner. The STB follows the design of~\cite{liu2021swin}, as illustrated in Fig.~\ref{fig:modules} (d). It applies the standard multi-head self-attention mechanism~\cite{vaswani2017attention} to locally shifted windows. First, the input features are reshaped from $\mathbb{R}^{H\times W\times C}$ to $\mathbb{R}^{\frac{HW}{M^2}\times M^2\times C}$ by dividing the features into $\frac{HW}{M^2}$ non-overlapping local windows of shape $M\times M$. In our case, we set $M=8$. Then, self-attention mechanism is applied to the features $F\in \mathbb{R}^{M^2\times C}$ in each local window. The \textit{query}, \textit{key}, and \textit{value} matrices $Q$, $K$, and $V$ are calculated as follows:
\begin{equation}
    Q = FP_Q,\quad K = FP_K,\quad V = FP_V,
\end{equation}
where $P_Q$, $P_K$, and $P_V$ are shared projection matrices across local windows, and $Q$, $K$, and $V$ are projected features with shape $\mathbb{R}^{M^2\times d}$. The process of self-attention is described as follows:
\begin{equation}
    \text{Attention}\left(Q,K,V\right) = \text{SoftMax}\left( \frac{QK^{T}}{\sqrt{d}} + B \right)V,
\end{equation}
where $B$ denotes the learnable relative positional encoding. Besides, multi-head mechanism is adopted to the self-attention (MSA), \ie, performing self-attention in parallel on the channel dimensions. We set the number of heads to 6 and the total number of channels to 174. Along with a multi-layer perception (MLP) and LayerNorm operation, the whole process of STB is as follows:
\begin{align}
    F &= MSA(LN(F)) + F,\\
    F &= MLP(LN(F)) + F.
\end{align}
Besides, shifted window partition is alternated between blocks to achieve cross-window connections, where the shift size is half of the window size $M$.

\begin{figure}[!t]
\centering
    \includegraphics[width=.8\linewidth]{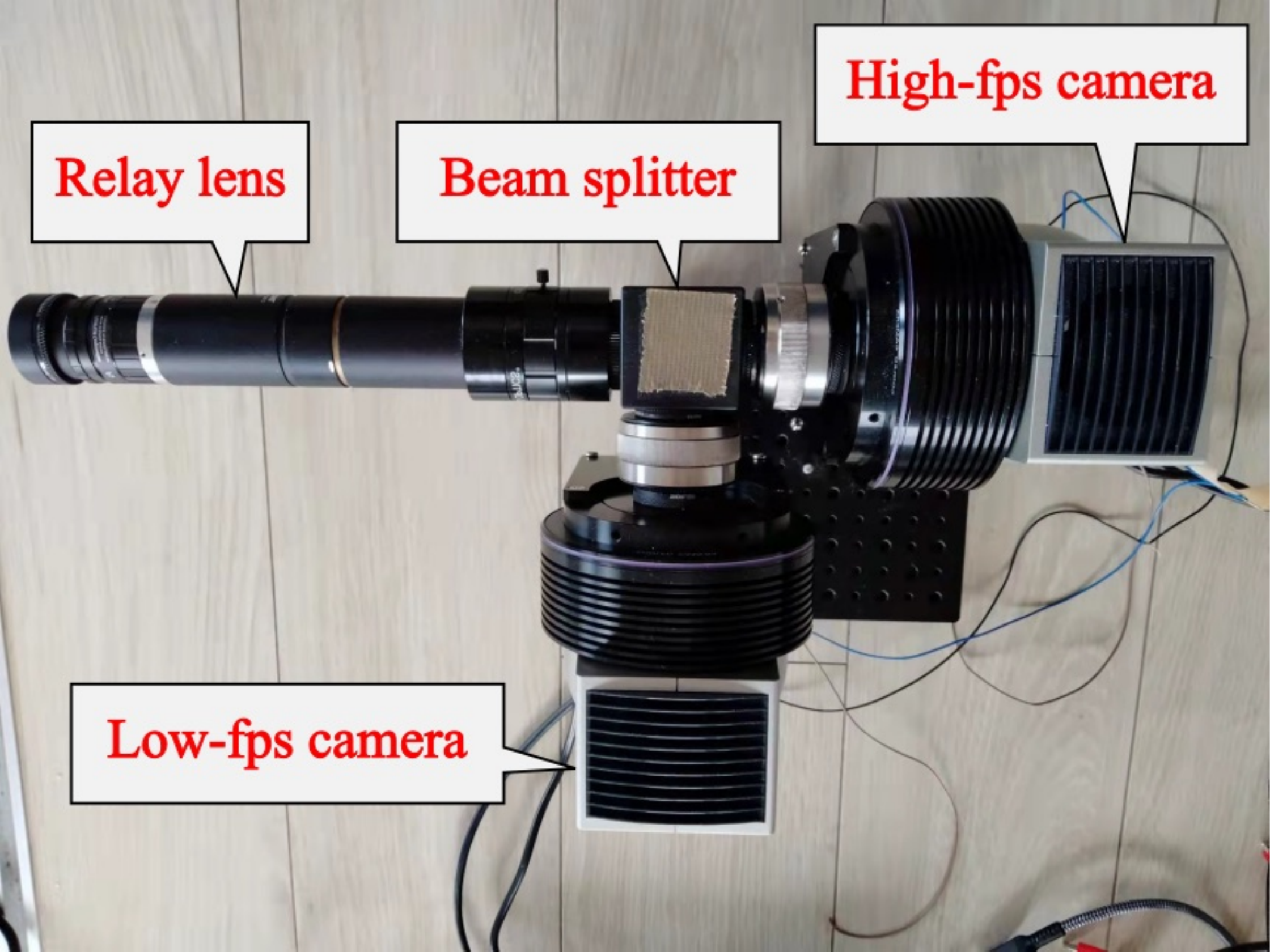}
    \caption{\textbf{Hardware of the proposed hybrid camera system.}}
    \label{fig:hardware}
\end{figure}

\begin{figure*}[!t]
\centering
    \includegraphics[width=\linewidth]{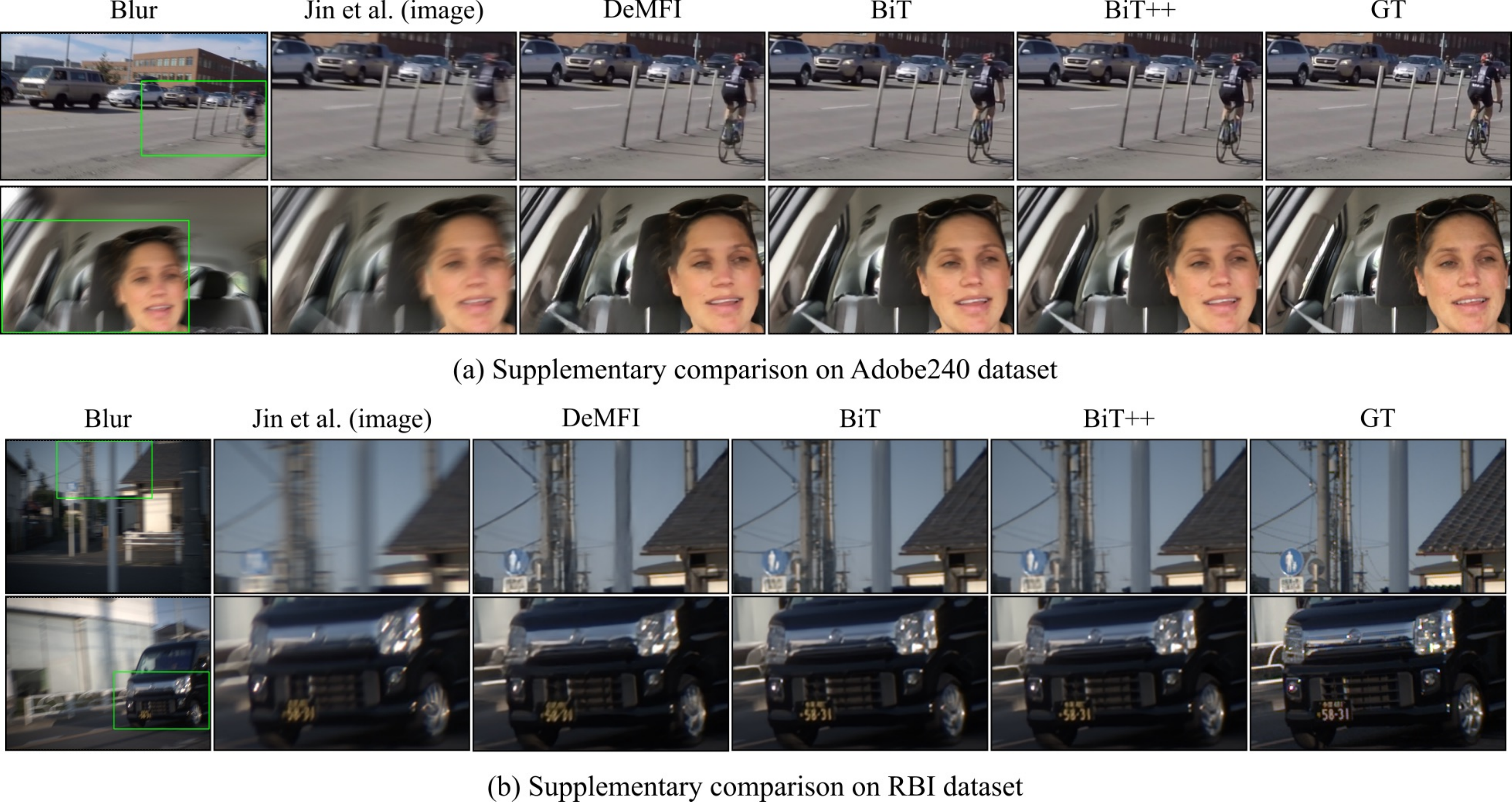}
    \caption{\textbf{Supplementary comparison on Adobe240~\cite{shen2020video} and the real-world RBI dataset.}}
    \label{fig:additional}
\end{figure*}

\begin{figure*}[!t]
\centering
    \includegraphics[width=\linewidth]{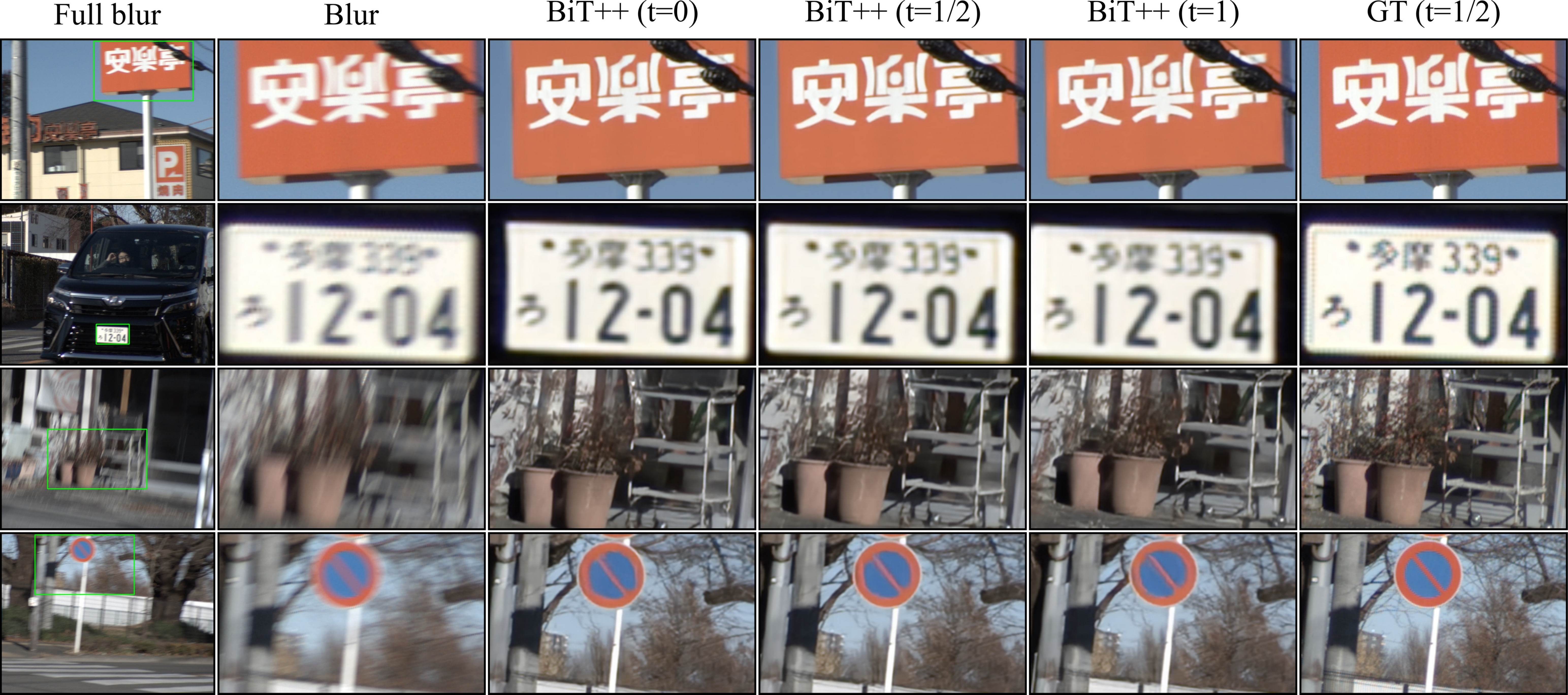}
    \caption{\textbf{Cross-validation on BSD dataset.}}
    \label{fig:third}
\end{figure*}

\begin{table}[!t]
	\caption{\textbf{Configurations of RBI dataset.}}
	\label{table:configuration}
	\centering
	\setlength{\tabcolsep}{12pt}
	\begin{tabular}{lcc}
    \toprule
    &  Train & Test\\
    \midrule
    Video pairs & 50 & 5\\
    Blur frame-rate & 25 & 25\\
    Sharp frame-rate & 500& 500\\
    Blur exposure time & \multicolumn{2}{c}{18ms} \\
    Sharp exposure time & \multicolumn{2}{c}{$\approx$2ms} \\
    Total blurred frames & 1250& 125\\
    Total sharp frames & 25000& 2500\\
    Resolution & \multicolumn{2}{c}{$640\times480$}\\
    Camera& \multicolumn{2}{c}{BITRAN CS-700C}\\
    \bottomrule
  \end{tabular}
\end{table}

\begin{table*}[!t]
\setlength{\tabcolsep}{12pt}
  \caption{\textbf{Effect of selected indices for DTS strategy.} The modifications are based on BiT.}
  \label{table:dts_ablation}
  \centering
  \begin{tabular}{lccccc}
    \toprule
    & t=1/2 \& t=1/2 & t=3/8 \& t=5/8 & t=1/4 \& t=3/4 & t=1/8 \& t=7/8 & t=0 \& t=1\\
    \midrule
    PSNR $\uparrow$ & 29.70 & 29.79 & 29.86 & 29.91 & 29.90\\
    SSIM $\uparrow$ & 0.894 & 0.895 & 0.897 & 0.898 & 0.900\\
    \bottomrule
  \end{tabular}
\end{table*}

\begin{table}[ht]
\setlength{\tabcolsep}{16pt}
  \caption{\textbf{Additional ablation for DTS.}}
  \label{tab:dts_t}
  \centering
  \begin{tabular}{lcc}
    \toprule
    & PSNR $\uparrow$ & SSIM $\uparrow$\\
    \midrule
    BiT (DTS w/ t) & 29.24 & 0.891 \\
    BiT & 29.90 & 0.900 \\
    \bottomrule
  \end{tabular}
\end{table}

\begin{table}[ht]
\setlength{\tabcolsep}{4pt}
  \caption{\textbf{Additional ablation for TSE.}}
  \label{tab:tse_reconstruct}
  \centering
  \begin{tabular}{lccc}
    \toprule
    & PSNR $\uparrow$ & SSIM $\uparrow$ & Parameters [M] $\downarrow$\\
    \midrule
    BiT+ (larger) & 30.12 & 0.902 & 12.049 \\
    BiT++ & 30.45 & 0.908 & 11.345 \\
    \bottomrule
  \end{tabular}
\end{table}

\begin{table}[!t]
\setlength{\tabcolsep}{5pt}
  \caption{\textbf{Ablation study of t encoding scheme.}}
  \label{table:t_encoding}
  \centering
  \begin{tabular}{lcccc}
    \toprule
    & BiT (freq.) & BiT & BiT (freq.) & BiT\\
    \midrule
    Dataset & Adobe240 & Adobe240 & RBI & RBI\\
    \midrule
    PSNR $\uparrow$ & 34.27 & 34.34 & 29.85 & 29.90 \\
    SSIM $\uparrow$ & 0.948 & 0.948 & 0.897 & 0.900 \\
    \bottomrule
  \end{tabular}
\end{table}

\begin{table}[!t]
\setlength{\tabcolsep}{6pt}
  \caption{\textbf{Effect of pretraining from Adobe240 to RBI.} The metrics are calculated only using middle predicted results ($t=0.5$).}
  \label{table:pretraining}
  \centering
  \begin{tabular}{lcccc}
    \toprule
    & BiT & Pre-BiT & BiT++ & Pre-BiT++\\
    \midrule
    PSNR $\uparrow$ & 29.90 & 30.79 & 30.45 & 31.32\\
    SSIM $\uparrow$ & 0.900 & 0.916 & 0.908 & 0.922\\
    \bottomrule
  \end{tabular}
\end{table}

\begin{table}[!t]
	\caption{\textbf{Comparison with Jin~\etal~\cite{jin2018learning} that takes single blurred image as input on synthetic dataset Adobe240 and our real-world dataset RBI.} \bestscore{Red} denotes the best performance, and \secondscore{blue} denotes the second best performance.}
	\label{table:comparison_additional}
	\centering
	\setlength{\tabcolsep}{4pt}
	\begin{tabular}{lcccc}
		\toprule
		& \multicolumn{2}{c}{Adobe240} & \multicolumn{2}{c}{RBI} \\
		\cmidrule(r){2-3}
		\cmidrule(r){4-5}
		& PSNR $\uparrow$ & SSIM $\uparrow$ & PSNR $\uparrow$ & SSIM $\uparrow$\\
		\midrule
		Jin~\etal~\cite{jin2018learning} & 25.03 & 0.776 & 25.27 & 0.814\\ 
		\midrule
		BiT & \secondscore{34.34} & \secondscore{0.948} & \secondscore{29.90} & \secondscore{0.900}\\
		BiT++ & \bestscore{34.97} & \bestscore{0.954} & \bestscore{30.45} & \bestscore{0.908}\\
		\bottomrule
	\end{tabular}
\end{table}

\paragraph{Real-world blur interpolation dataset (RBI).} The actual diagram of our hybrid camera is illustrated in the Fig.~\ref{fig:hardware}. In addition, the detailed configurations are shown in Table~\ref{table:configuration}. As for geometric alignment, the two cameras are first mechanically aligned assisted with collimated laser beams. Later, a homography correction using standard checker pattern is conducted, so as to reduce the alignment error to less than one pixel. Lens distortion will occur when two lenses are behind the beam-splitter, thus we put the lens in front (only one lens). Even with any distortion, the two cameras are identical, so the effect on learning is limited. There is no post-processing, such as flow-based alignment, but only homography. This real-world dataset can be applied to multiple applications, such as image/video deblurring, blur interpolation, and blur synthesis~\cite{brooks2019learning}. By simply modifying the parameters of the hardware, we can obtain a richer and more diverse dataset in the future.

\section{Supplementary results}

\paragraph{Effect of selected indices for DTS strategy.} 
We present an ablation study of the indices selected for the DTS strategy in Table~\ref{table:dts_ablation}. t=0 \& t=1 represents the default setting for DTS strategy. These ablation studies are conducted on RBI dataset. Experimental results show that it is better to select sharp frames closer to the dual-end of the exposure time as supervision.

\paragraph{Additional ablation for DTS.} 
We add an ablation study of supervising dual-end by going through $F_M$ with t as input, namely ``BiT (DTS w/ t)'', as shown in Table~\ref{tab:dts_t}. The results support that DTS can make the form of shared features more conducive to arbitrary interpolation.

\paragraph{Additional ablation for TSE.} 
In contrast to BiT/BiT+, BiT++ has a larger reconstruction layer with 11.345M parameters, which is a mere 0.67\% increase over the 11.270M parameters in BiT/BiT+. We train a larger BiT+ network with 12.049M parameters by increasing the output channels of $F_M$ without TSE, as shown in Table~\ref{tab:tse_reconstruct}. The results support that TSE actually brings extra information more than the effect of more parameters.

\paragraph{Ablation study of t encoding scheme.}
We show the comparison to commonly used frequency encoding in Table~\ref{table:t_encoding}. As mentioned in the manuscript, we find that simple encoding by concatenating feature channels can provide good enough performance, even slightly better than the widely used frequency encoding.

\paragraph{Pretraining using Adobe240.}
We show the effect of pretraining from Adobe240~\cite{shen2020blurry,shen2020video} to our RBI in Table~\ref{table:pretraining}. The BiT initialized with checkpoints of the BiT trained on Adobe240 is denoted as Pre-BiT, and the corresponding full model with temporal symmetry is denoted as Pre-BiT++. The results show that pretraining with Adobe240 can bring benefits to the model. Although the data of Adobe240 is synthetic, its scene diversity is useful to the model.

\paragraph{Supplementary comparison results.} 
Due to the directional blurring problem~\cite{zhong2022animation}, it is very unlikely that the method with one blurred image as input produces results with the same decomposition order as the ground-truth sequence. The poor quantitative performance of Jin~\etal~\cite{jin2018learning} in Table~\ref{table:comparison_additional} prove this point. Besides, we supplement the visual results of Jin~\etal~\cite{jin2018learning} and DeMFI~\cite{oh2021demfi} on both Adobe240=\cite{shen2020video} and the real-world RBI dataset to further validate the superior performance of our method, as illustrated in Fig.~\ref{fig:additional}.

\paragraph{Third-part data validation.} To further validate the robustness of the model trained on our real-world dataset, RBI, we test our model on the third-party data from BSD~\cite{zhong2020efficient}. The qualitative results are shown in Fig.~\ref{fig:third}. Sharp motion sequence are successfully rendered out of the blurred images. This again highlights the necessity and importance of proposing a real dataset.

\end{document}